\documentclass[letterpaper]{article} 
\usepackage{aaai24}  
\usepackage{times}  
\usepackage{helvet}  
\usepackage{courier}  
\usepackage[hyphens]{url}  
\usepackage{graphicx} 
\usepackage{natbib}  
\usepackage{caption} 
\usepackage{algorithm}
\usepackage{algorithmic}

\usepackage{newfloat}
\usepackage{listings}
\DeclareCaptionStyle{ruled}{labelfont=normalfont,labelsep=colon,strut=off} 
\floatstyle{ruled}
\newfloat{listing}{tb}{lst}{}
\floatname{listing}{Listing}
\usepackage{makecell}
\usepackage{subcaption}
\usepackage{booktabs}       
\usepackage{multirow}
\usepackage{amssymb}
\usepackage{amsmath}
\usepackage[dvipsnames, svgnames, x11names]{xcolor}
\usepackage{colortbl}

\usepackage{cleveref}

\title{Static-Dynamic Class-level Perception Consistency in Video Semantic
Segmentation}
\author{Zhigang Cen, Ningyan Guo, Wenjing Xu, Zhiyong Feng, Danlan Huang}

\usepackage{bibentry}

\begin{document}

\maketitle

\begin{abstract}
Video semantic segmentation(VSS) has been widely employed in lots of fields, such as simultaneous localization and mapping, autonomous driving and surveillance. Its core challenge is how to leverage temporal information to achieve better segmentation. Previous efforts have primarily focused on pixel-level static-dynamic contexts matching, utilizing techniques such as optical flow and attention mechanisms.
Instead, this paper rethinks static-dynamic contexts at the class level and proposes a novel static-dynamic class-level perceptual consistency (SD-CPC) framework. In this framework, we propose multivariate class prototype with contrastive learning and a static-dynamic semantic alignment module. The former provides class-level constraints for the model, obtaining personalized inter-class features and diversified intra-class features. The latter first establishes intra-frame spatial multi-scale and multi-level correlations to achieve static semantic alignment. Then, based on cross-frame static perceptual differences, it performs two-stage cross-frame selective  aggregation to achieve dynamic semantic alignment. Meanwhile, we propose a window-based attention map calculation method that leverages the sparsity of attention points during cross-frame aggregation to reduce computation cost. Extensive experiments on VSPW and Cityscapes datasets show that the proposed approach outperforms state-of-the-art methods. Our implementation will be open-sourced on GitHub.
\end{abstract}

\section{Introduction}\label{sec:intro}
Semantic segmentation aims to assign a semantic label for each pixel of the images, which is widely employed in lots of fields, such as simultaneous localization and mapping, autonomous driving and surveillance \cite{slam.semantic1,motion-state,UAVformer,atuonomous,surveillance}. Benefiting from the abundant datasets \cite{ade20k,cityscapes} of image semantic segmentation (ISS) and the powerful feature extraction of the deep neural networks \cite{ dilated_conv,tensor,pyramid,vit,UperNet_Netwarp,efficientvit}, ISS has made significant progress during the past few years \cite{efficientvit,flow-propagation, fcn_vss, xie2021segformer, mobilevit, swin_at_length4,efficientnet}. However, the real world comprises a sequence of video frames rather than a single image. Consequently, the video semantic segmentation (VSS) has gained great attention in recent years, but it also encounters new challenges.

\begin{figure}[t]
  \centering
  \begin{subfigure}{\columnwidth}
    \includegraphics[width=\columnwidth]{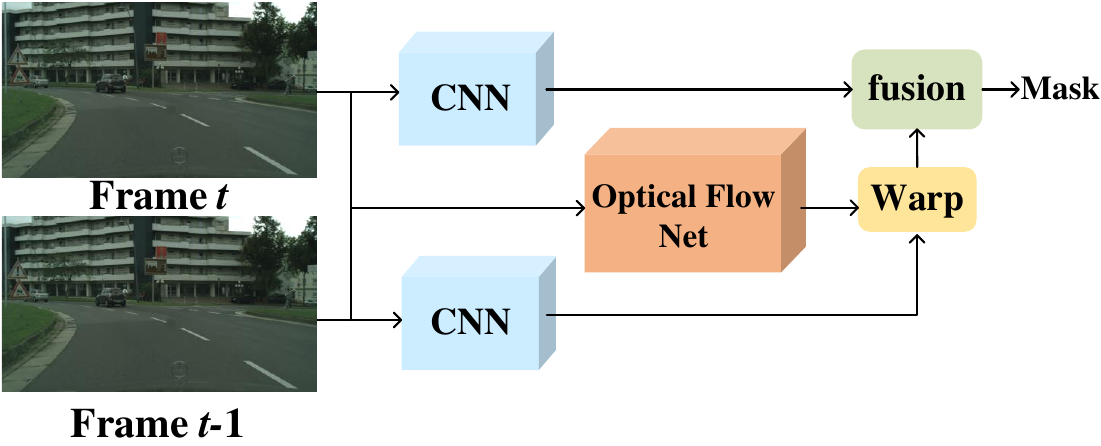}
    \caption{Direct Method}
    \label{fig:existing_method_1}
  \end{subfigure}
  \begin{subfigure}{\columnwidth}
    \includegraphics[width=\columnwidth]{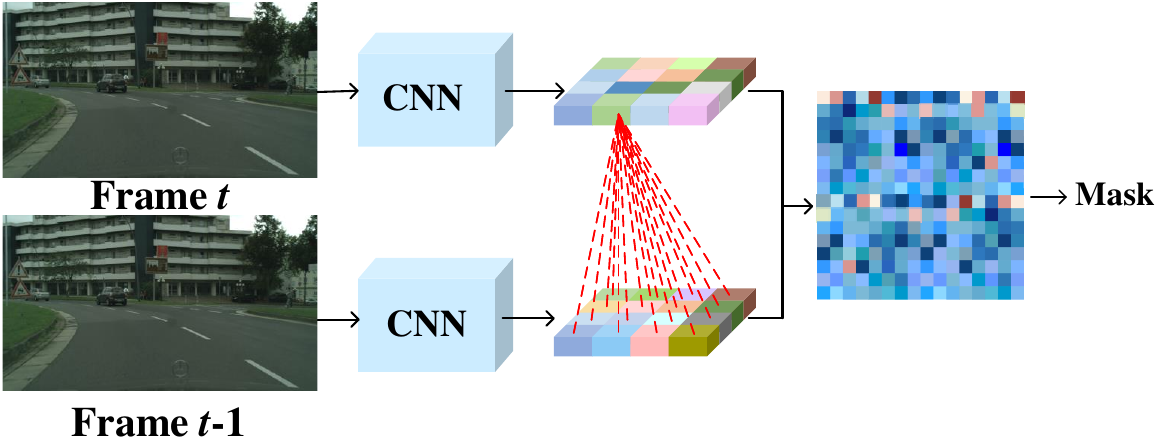}
    \caption{Indirect Method}
    \label{fig:existing_method_2}
  \end{subfigure}
  \begin{subfigure}{\columnwidth}
    \includegraphics[width=\columnwidth]{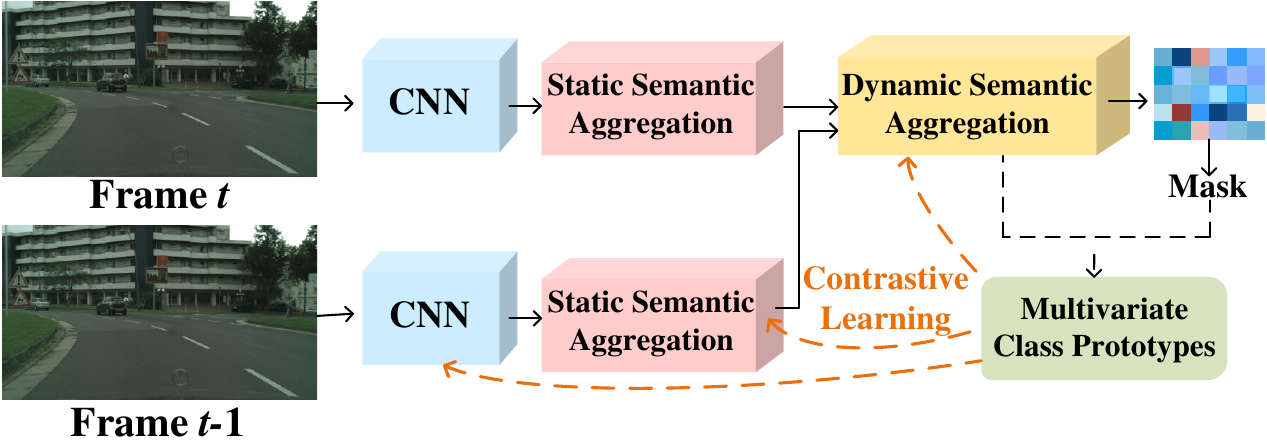}
    \caption{The Proposed Method }
    \label{fig:existing_method_our}
  \end{subfigure}
\caption{Comparison of the different methods. (a) The direct methods explicitly distort features based on pre-trained optical flow networks, resulting in inconsistent information. (b) The indirect methods model the relationship between all pixels with the attention mechanism, leading to extremely high computation cost. (c) The proposed method models the static-dynamic spatio-temporal associations at the category level, achieving more efficient and accurate results.}
\label{fig:existing_method}
\end{figure}

Compared to the ISS, the core of the VSS is how to effectively leverage spatio-temporal contextual information. It is widely accepted that contextual information can be categorized into static and dynamic contexts \cite{eventfultransformer, domain-opticalflow, flow-propagation, motion-state, perceptual-consistency, coarse, accel, dvsnet, zhu2017deep}. The former refers to the contexts within a single video frame or the contexts of consistent content between consecutive frames, encompassing more detailed semantic region information. The latter refers to cross-frame motion information and spatio-temporal associations, facilitating the matching of semantic regions across frames and reducing segmentation uncertainty. Many existing works \cite{change-resolutions,coarse,domain-opticalflow,flow-propagation,interfuture-fusion,motion-state,mining-relation, perceptual-consistency,mask-propagation} leveraging cross-frame associations have achieved impressive results and can be summarized into two main categories, $i.e.$, direct methods and indirect methods, as illustrated in Figure \ref{fig:existing_method}. Direct methods explore spatio-temporal correlation by warping the features from the previous frame to the current frame using an additional optical flow network, thereby improving segmentation performance. However, due to the scarcity of datasets that have both segmentation and optical flow annotations, the end-to-end optimization is challenging. Additionally, the method is susceptible to occlusions and fast-moving objects, which degrade segmentation quality. Indirect methods utilize attention mechanisms \cite{transformer} to implicitly capture the dependencies of all the pixels within intra-frames and cross-frames. However, this method results in the high computational complexity due to computing the correlation matrix for all pixels between frames.

To address the above problems, inspired by \cite{multispectral, motion-state, perceptual-consistency}, we rethink the static and dynamic contexts in VSS from the perspective of class-level perception consistency. For static contexts, compared to isolated pixels, pixels belonging to the same semantic category exhibit a regional distribution within a frame and their semantic features are highly similar in the class feature space. For dynamic contexts, compared to pixel-level associations, the categories between adjacent frames maintain higher perceptual similarity in terms of category types, semantic features, spatial locations, and motion patterns. Overall, at the class level, image elements can be simplified to facilitate easier implementation of spatio-temporal associations. This simplification allows the model to focus more on extracting and matching category features, rather than on learning irrelevant pixel details for the segmentation task. Therefore, we propose a novel framework called the static-dynamic class-level perception consistency (SD-CPC). Specifically, we propose the multivariate class prototype with contrastive learning (MCP-CL) to constrain the similarity of inter-class and intra-class features. This ensures the separability of class features. Subsequently, based on class-level perceptual consistency, we propose a static-dynamic semantic alignment module composed of the static semantic efficient aggregation module (SSEA) and the dynamic semantic selective aggregation module (DSSA). SSEA models the spatial relationships in each frame at multi-scale and multi-level, thereby achieving static semantic alignment. Subsequently, after interleaving features output by SSEA from different frames, DSSA performs convolutions on these features to capture cross-frame perceptual differences. Based on these perceptual differences, DSSA conducts a two-stage selective aggregation of adjacent frame pixels from coarse to fine, achieving dynamic semantic alignment. In this process, as we only aggregate partial regions from adjacent frames, we reconstruct the query ($Q$), key ($K$), value ($V$) matrices in a windowed manner, and compute the attention map through Hadamard product, thereby reducing the computation cost.

The various parts of the framework are tightly coupled, making the entire paradigm ingenious. SSEA effectively captures static semantics and long-range relationships, providing reliable static perceptual differences for DSSA to achieve cross-frame selective aggregation. DSSA enhances current frame features through capturing motion information, allowing SSEA to avoid the need of complex designs in ISS. MCP-CL provides the model with class-level learning constraints and enhances the representational capacity of class features through a multivariate approach.

Overall, our contributions are as follows:
\begin{itemize}
\item From the perspective of class-level perceptual consistency, we propose a novel VSS framework to achieve a better trade-off between performance and efficiency.
\item We design a static-dynamic semantic alignment module to explore class-level spatio-temporal relationships. And, we propose a window-based attention map calculation method that leverages the sparsity of attention points during cross-frame aggregation to reduce computation cost.
\item We propose multivariate class prototype with contrastive learning, which not only provides class-level perceptual constraints but also enhances the model's representation capabilities through a multivariate approach.
\end{itemize}

\section{Related Works}
\subsection{Direct Methods}
Direct methods \cite{accel,reconstruction,change-resolutions,dvsnet,UperNet_Netwarp,zhu2017deep,ding2020every,zhu2019improving} typically distort features from previous frames to the current frame based on optical flow obtained from pre-trained optical flow networks, achieving spatio-temporal consistency. Accel \cite{accel} propagates detailed information in reference branch and conducts feature warping via optical flow. IFR\cite{reconstruction} reconstructs the current frame features by obtaining class prototypes from the reference frame, which improves training efficiency and segmentation performance. AR-Seg \cite{change-resolutions} uses different resolutions for key frames and non-key frames, and distorts features via motion vectors, thereby reducing computation costs. DVSNet \cite{dvsnet} divides the current frame into different regions and performs different operations based on the differences among these regions, thereby achieving a balance between performance and efficiency. Although these methods can capture spatio-temporal information between frames and offer good interpretability, they still encounter challenges such as difficulties in end-to-end optimization, susceptibility to error propagation, and vulnerability to environmental influences.

\subsection{Indirect Methods}
Indirect methods \cite{coarse,ETC,motion-state,multispectral,mining-relation,k-net,uvid-net,temporal-memory, perceptual-consistency,mining_weak} employ attention mechanisms to compute a relationship matrix, replacing optical flow, thereby utilizing cross-frame correlations and implicitly aligning features. For instance, CFFM-VSS \cite{coarse} employs different convolution and pooling for different moment frames, and mines temporal features through multi-head non-self attention. MRCFA \cite{mining-relation} achieves better aggregation of temporal information by exploring the relationships between cross-frame affinities. ETC \cite{ETC} proposes a time knowledge distillation method to reduce the performance gap between models. MSAF \cite{motion-state} aligns static and dynamic semantics through motion and status branches, respectively, and links pixel-level descriptors with region-level descriptors using semantic assignment. MVSS \cite{multispectral} reduces the computation costs of the attention between multi-modalities and multi-frames by class prototypes. However, these methods have not fully exploited the abundant redundancy of cross-frame information, focusing solely on pixel-level implicit correlations while neglecting class-level constraints.

\section{Methodology}
\begin{figure*}[!t]
  \centering
  \includegraphics[width=\linewidth]{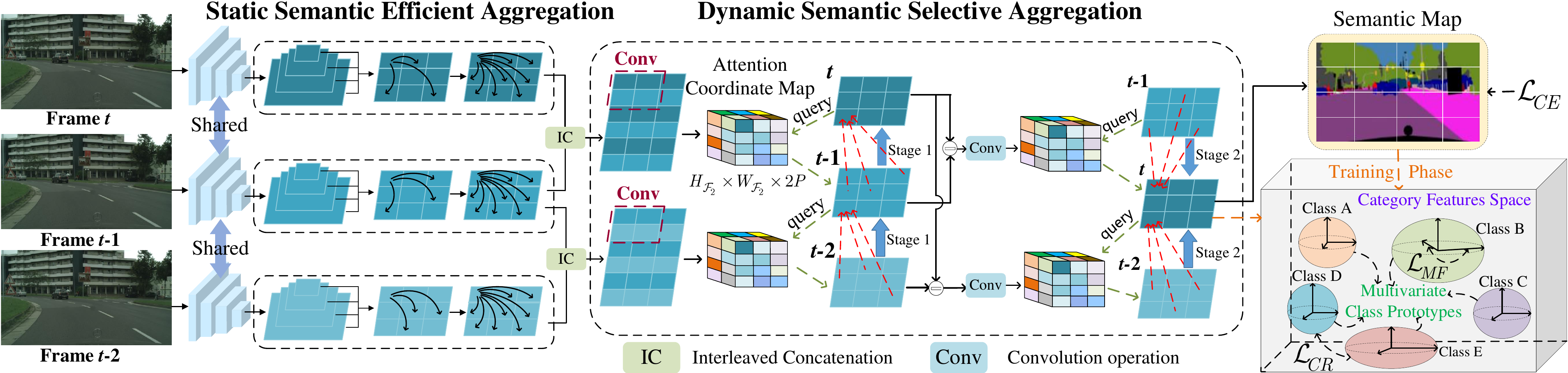}
  \caption{Framework of the proposed SD-CPC framework. First, we model the spatial relationship of pixel features extracted by the backbone at multi-scale and multi-level, achieving static semantic alignment. Then, based on the cross-frame static semantic differences , we conduct the two-stage dynamic semantic selective aggregation to achieve dynamic semantic alignment. During training, we obtain multivariate class prototypes based on the prediction results and output features, and then combine them with contrastive learning to realize class-level constraints and improve the model's representation capability.}
  \label{fig:framework}
\end{figure*}

In this section, we will introduce each component of the framework. The framework is illustrated in Figure \ref{fig:framework}.

\subsection{Static-Dynamic Semantic Alignment}\label{sec:3.2}

To extract feature for each frame, we use MiT-B1 \cite{xie2021segformer} as the feature extractor. Feature extractor has four stages to encode features from different scales, named as {$\mathcal F_s$}.
For an input image with height $H$ and width $W$, the feature map corresponding to the first stage encoding is $\mathcal{F}_1$ with the size of $H_{\mathcal{F}_1} \times W_{\mathcal{F}_1} \times C_{\mathcal{F}_1}$, where $H_{\mathcal{F}_1}$, $W_{\mathcal{F}_1}$, and $C_{\mathcal{F}_1}$ represent the height, width, and number of channels of the feature map, respectively. In each following stage, the dimensions of the feature map are halved, while the number of feature channels is increased.

\noindent
{\bf Static Semantic Efficient Aggregation}. Static semantic encompasses detailed semantic information of the current frame, and serves as the foundation for cross-frame selective aggregation of dynamic semantics. Previous studies \cite{pyramid, xie2021segformer, dilated_conv, DeepLabv3+} have demonstrated that multi-scale and global receptive fields are crucial for semantic segmentation. Moreover, low-level features tend to have larger sizes and contain more detailed information, while high-level features usually have smaller sizes and richer semantic content.

Therefore, in SSEA, we first conduct multi-scale fusion of features $\mathcal{F}_1$, $\mathcal{F}_2$, $\mathcal{F}_3$, $\mathcal{F}_4$ from different stages of the backbone to enhance feature representations. To establish multi-level spatial correlations with low computational cost, we then combine deformable convolution (DCN) \cite{dcn} and linear attention (LA) \cite{LA}. The local deformable receptive fields provided by DCN refine the global associations of LA, mitigating LA's poor focusing performance. LA can provide DCN with a larger receptive field range for selecting deformable convolution regions. Compared to the vanilla softmax attention mechanism($\mathcal{O}(N^2C)$, where $N=H_{\mathcal{F}_2} \times W_{\mathcal{F}_2}, C=C_{\mathcal{F}_2}, N\gg C$), this simple yet effective method retains the low computation cost advantage of LA while also mitigating its poor performance in long-distance modeling (see Table \ref{tab:abl_SD-CPC} for experimental results). The computation cost of the module is $\mathcal{O}(NDC+NC^2)$, where $D$ is the number of aggregation points. The formal definition of the SSEA is as follows:
\begin{equation}
    {\mathcal S}= \textrm{LA}(\textrm{DCN}([\textrm{DS}({\mathcal F}_1^{'})\oplus {\mathcal F}_2\oplus\textrm{US}({\mathcal F}_3^{'})\oplus \textrm{US}({\mathcal F}_4^{'})])), \label{eq:SSEA}
\end{equation}
where  $\oplus$ denotes the feature concatenation operation, US denotes up-sampling, and DS denotes down-sampling. 

\noindent
{\bf Dynamic Semantic Selective Aggregation}. Due to the strong correlation between adjacent frames in terms of category features, semantic categories, category spatial distributions, and motion patterns, this provides two important benefits: (1) Capturing dynamic associations between frames helps reduce the uncertainty of perception. (2) Considering the high similarity in category perception between adjacent frames, there exists significant redundant information across frames. Therefore, the current frame only needs to selectively aggregate partial pixel information from the previous frame to achieve dynamic semantic alignment. Additionally, as the time interval increases, the range of pixel changes and related areas between frames also expands. This implies the need for a larger receptive field to capture global information. Therefore, we conduct a two-stage cross-frame selective cross-attention on dynamic semantics from coarse to fine, leveraging static perceptual differences.

Specifically, in the first stage, we interleave $\mathcal{S}^{t-1}$ and $\mathcal{S}^{t-2}$ along rows (or columns), and the results are fed into two convolutional layers. This process generates an attention coordinate map of size $H_{\mathcal{F}_2}\times W_{\mathcal{F}_2} \times 2P$ based on the perceptual differences of the interleaved rows (or columns). The coordinate map records the coordinates of $P$ pixels in $\mathcal{S}^{t-2}$ that are of interest to each pixel in $\mathcal{S}^{t-1}$. Based on these coordinates, each pixel in $\mathcal{S}^{t-1}$ selectively attends to $P$ pixels in $\mathcal{S}^{t-2}$ through cross-frame selective cross-attention mechanism to obtain $\mathcal{D}^{t-1}$. This achieves the selective aggregation of spatio-temporal information. Then, we repeat the aforementioned steps with $\mathcal{S}^{t}$ and $\mathcal{D}^{t-1}$ to obtain $\mathcal{D}_{\textrm{coase}}^{t}$. This progressive aggregation method not only gradually aligns multiple frames but also allows the current frame to have a greater receptive field for frames with longer time intervals. Since the current frame (time slot $t$) and the reference frame (time slot $t-1$ and $t-2$) have now established a rough spatio-temporal association, the second stage aims to refine this association. In the second stage, we directly subtract the reference frame from the target frame to obtain pixel-wise perceptual differences and then apply convolution to generate the attention coordinate map.
Then, $\mathcal{D}_{\textrm{fine}}^t$ is obtained by repeating the aforementioned steps. This two-stage aggregation method effectively establishes multi-scale spatio-temporal correlations from coarse to fine.


We use $\mathcal{S}_{t-1}$ and $\mathcal{S}_{t-2}$ as examples to describe the cross-frame selective cross-attention mechanism. First, $\mathcal{S}_{t-1}$ is fed into a multi-layer perceptron (MLP) to obtain $Q$, while $\mathcal{S}_{t-2}$ is fed into a MLP to obtain $K$ and $V$. Subsequently, based pn the attention coordinate map, we extract $N \times P$ pixel features from $K$ and $V$, which are then partitioned into $N$ windows of size $\sqrt{P} \times \sqrt{P}$. Each window corresponds to the region of interest for each pixel in $\mathcal{S}_{t-1}$ within $\mathcal{S}_{t-2}$. $Q$ is expanded to match the $K $($V$) dimensions. Finally, we conduct the attention mechanism within each window in parallel, thereby achieving dynamic semantic aggregation. The computation process for each window is as follows:
\begin{equation}
O_{w} = \sum_{p=1}^{P}\frac{\sum_{c=1}^{C}Q^{p,c}_{w}\odot K^{p,c}_{w}}{\sum_{p=1}^{P}\sum_{c=1}^{C}Q^{p,c}_{w}\odot K^{p,c}_{w}}V_w^{p,(\cdot)},
\label{eq:caoca}
\end{equation}
where $O_w$ represents the aggregation result of the $w$-th window, $ Q_w^{p,c}/K_w^{p,c} $ denotes the $c$-th channel of the $p$-th element in the $w$-th window, and $\odot$ signifies the Hadamard product. This method leverages the sparsity of cross-frame attention points, reducing the computation cost from $\mathcal{O}(N^2)$ to $\mathcal{O}(NP)$, where $N \gg P$.

\noindent
{\bf Difference with DAT}. It is worth noting that while the proposed model and the deformable attention transformer (DAT) \cite{xia2022vision} both obtain irregular attention regions, there are several key differences between them. Firstly, the motivations are different. Our motivation is to capture spatio-temporal information and avoid redundant computation through static-dynamic class-level perception consistency. Therefore, each pixel in the current frame selects the region of interest from the previous frame. This is a filtering process of the original attention regions, rather than a shifting of attention points. In contrast, DAT aims to achieve a larger receptive field by translating rectangular aggregation areas into irregular regions through attention points offsets. Secondly, the designs are different. The proposed method determines the regions of interest based on static perceptual differences, whereas DAT directly conducts convolution on the surrounding area of the anchor point to obtain attention offsets. Therefore, the proposed method has better interpretability and reliability. Furthermore, the proposed method employs a cross-frame cross-attention mechanism, leveraging the sparsity of attention points to improve efficiency, while DAT employs an intra-frame self-attention mechanism. Finally, the focus on dimensions differs. the proposed method achieves spatio-temporal multi-scale selective aggregation through a two-stage process, while the DAT model performs single-stage spatial dynamic aggregation within each frame.

\noindent
{\bf Complexity analysis}. Following the setting in \cite{motion-state}, We will analyze the complexity under the premise of ignoring the impact of scaled dot-product and multi-head on reducing the amount of computation. Therefore, the computation cost that constructs the relation between one pixel with all other pixels in spatio-temporal dimension is $\mathcal{O}(N^2C^2)$. And, the total computation cost of Transformer \cite{transformer} between frame $t$ and $t-1$ as well as $t-2$ is $\mathcal{O}(N^3C^3)$. Therefore, the whole computation cost of vanilla Transformer is $\mathcal{O}(3N^3C^3)$.

In the proposed method, the computation cost of cross-frame selective cross-attention mechanism can be regarded as $\mathcal{O}(NPC^2)$. Therefore, the computation cost of SSEA is $\mathcal{O}(3(NDC+NC^2))$, the computation cost of DSSA is $\mathcal{O}(3NDC+4NPC^2)$, and the whole computation cost of the model is $\mathcal{O}(6NDC+3NC^2+4NPC^2)$, where usually $N>>D$ (or $C,P$) and $C>P\approx D$. Obviously, compared with $\mathcal{O}(N^3C^3)$, the complexity of the proposed method increases linearly with respect to $N$, so the proposed method is more efficient than vanilla Transformer.

\subsection{Mlutivarite Class Prototypes with Contrastive Learning}
Class prototypes represent the feature centroids of each semantic category. During training, Existing methods \cite{multispectral,motion-state,reconstruction,interfuture-fusion} iteratively update class prototypes to assign semantic labels. However, in the iterative process, the inaccuracy in class prototype calculation and the incomplete class coverage increase the difficulty of training. Moreover, class prototypes constructed from single features are overly simplistic and are unable to withstand variations caused by environmental factors (e.g., lighting) and individual differences. Inspired by how humans recognize objects by comparing multiple aspects, as well as existing works \cite{cl,cl2}, we propose the MCP-CL. This method computes class prototypes only among correctly predicted pixels and utilizes contrastive learning to constrain class feature differences, thereby achieving class-level perceptual consistency. This method not only avoids the problem of incomplete class coverage but also provides stronger class-level constraints as the network segmentation accuracy improves. Furthermore, we enhance class representation capabilities through multivariate joint representation.

Specifically, we project $\mathcal{D}_{\textrm{fine}}^t$ to obtain the multivariate feature $\mathcal{M}^t \in \mathbb{R}^{M\times H_{\mathcal{F}_2} \times W_{\mathcal{F}_2} \times \frac{C}{M}}$, where $M$ represents the number of multivariate. Subsequently, each variate feature undergoes independent prediction, and the results are combined through joint decision-making to produce the final prediction, denoted as $\mathcal{S}^t$. The entire process is supervised learning based on minimizing the cross-entropy loss:
\begin{equation}
\mathcal{L}_{\text{CE}} = -\sum_{i=1}^{H \times W}\sum_{j=1}^{CLS}{\mathcal{G}^t_{i,cls}\log\mathcal{S}^t_{i,cls}}, \label{eq:entropy_loss}
\end{equation}
where $CLS$ is the number of class, ${\mathcal{G}}^t_{i,cls}$ is the real probabilities that $i$-th pixels belongs to $cls$-th class in $t$-th frame. 

During training, we take the intersection of $\mathcal{S}^t$ and $\mathcal{G}^t$ to obtain the prediction correct mask $\mathcal{G}_{\textrm{mask}}^t$, where the non-zero elements are $N_G$. Subsequently, we aggregate the features of pixels belonging to the same category in $\mathcal{M}^t$ according to $\mathcal{G}_{\textrm{mask}}^t$ to obtain the multivariate class prototype $\mathcal{P}^t_{cls}$. Formally, the calculation process can be formulated as follows:
\begin{equation}
\mathcal{P}^{t}_{cls} = \frac{\sum_{i=1}^{N_G}\mathcal{M}^{t} \cdot \mathbb{I}(\mathcal{G}_{\textrm{mask}}^t = cls)}{\sum_{i=1}^{N_G}\mathbb{I}(\mathcal{G}_{\textrm{mask}}^t = cls)},
\label{eq:class_prototype}
\end{equation}
where ${\mathbb{I}}$ is an indicator function. Subsequently, we employ contrastive learning to maximize the distance between the feature centroids of different classes, ensuring the separability of class-level features. For a query pixel $p$, the multivariate class prototype belonging to the same category is positive sample $p^+$, while multivariate class prototypes from different categories constitute the negative sample set $p^- \in \mathcal{N}$. Formally, the contrastive loss is as follows:
\begin{equation}
\begin{aligned}
\mathcal{L}_{\text{CR}} &= \frac{1}{M \times N_G}\sum_{m=1}^{M}\sum_{i=1}^{N_G}\mathcal{L}_{cl}(i,m), 
 \\
 \mathcal{L}_{cl}(i,m)&=\log\left(1 + \frac{\sum_{\mathbf{p}^{-}\in \mathcal{N}}\exp\left(\frac{\mathcal{M}^{t}_{i,m}\cdot\mathbf{p}^{-}}{\tau}\right)}{\exp\left(\frac{\mathcal{M}^{t}_{i,m}\cdot\mathbf{p}^{+}}{\tau}\right)}\right),
\end{aligned}
\label{eq:class prototype}
\end{equation}
where $\tau$ denotes the temperature parameter, $M_{i,m}^t$ represents the $m$-th multivariate feature of the $i$-th non-zero element in $\mathcal{G}_{mask}^{t}$. To enhance the model's representation capability, we constrain the similarity between each variate features in intra-class, the formal definition of which is as follows:
\begin{equation}
\mathcal{L}_{\text{MF}} = \frac{1}{N_G}\sum_{n=1}^{N_G}\frac{1}{C_M^2}\sum_{i=1}^M\sum_{j=i+1}^M\left(\mathcal{P}^t_{cls} \cdot \left(\mathcal{P}^t_{cls}\right)^\top\right)_{i,j}, 
\label{eq:PM}
\end{equation}
where $C_M^2$ represents the combinatorial number. This magnitude indicates the number of elements in the upper triangular part of the similarity matrix (${\mathcal{P}}^t_{cls} \cdot  ({\mathcal{P}}^t_{cls})^T$). The overall learning targets can be denoted as:

\begin{equation}
\mathcal{L}_{\text{total}} = \mathcal{L}_{\text{CE}} + \lambda_1\mathcal{L}_{\text{CR}} + \lambda_2\mathcal{L}_{\text{MF}}, \label{eq:total}
\end{equation}
where $\lambda_1$ and $\lambda_2$ are the weight parameters.

\section{Experiments}\label{sec: experiments}
{\bf Implementation details.} We conduct all experiments using two NVIDIA GeForce RTX 4090 GPUs.The backbones are the same as SegFormer\cite{xie2021segformer}. We use frames $t$-3 and $t$-6 as reference frames and set $P$=4, $D$=9, $M$=4. During training, we apply random resizing, flipping, cropping, and photometric distortion for data augmentation. The VSPW dataset \cite{vspw} crops each frame to 480$\times$480, while the Cityscapes dataset \cite{cityscapes} crops each frame to 512$\times$1024. We use the AdamW optimizer \cite{adamw} with a "poly" learning rate strategy, and set the initial learning rate to 0.00002. During testing, images are resized to 480$\times$853 for VSPW and 1024$\times$2048 for Cityscapes.

\begin{table*}[!t]
\caption{Performance comparisons with state-of-the-art methods on VSPW dataset. “-” indicates that data cannot be obtained. Note that our model achieves a better balance between accuracy, video consistency, and model complexity.}
\label{tab:vspw_sota}
\centering
\begin{tabular}{@{}c|c|c|c|c|c|c|c|c@{}}
\toprule
Methods    & Backbone   & mIoU$\uparrow$  & WIoU$\uparrow$  & mVC$_{8}\uparrow$  & mVC$_{16}\uparrow$    & GFLOPs$\downarrow$    & Params$\downarrow$ & FPS(f/s)$\uparrow$ \\
\midrule
SegFormer  & MiT-B1     & 36.5    & 58.8    & 84.7   & 79.9     & 26.6      & 13.8   & 68.67 \\
MRCFA   & MiT-B1     & 38.9    & 60.0    & 88.8   & 84.4     & -         & 16.2   & 30.90   \\
CFFM-VSS  & MiT-B1     & 38.5    & 60.0    & 88.6   & 84.1     & 103.1     & 15.5   & 31.06   \\
SD-CPC(ours) & MiT-B1  & {\bf 39.9}  & {\bf 60.8}  & {\bf 88.9}  & {\bf 84.7}  & 69.2 &15.0 & 32.54\\
\midrule
DeepLab3+ & Res-101    & 34.7    & 58.8    & 83.2   & 78.2     & 379.0     & 62.7   & - \\
UperNet   & Res-101    & 36.5    & 58.6    & 82.6   & 76.1     & 403.6     & 83.2   & -  \\
PSPNet & Res-101    & 36.5    & 58.1    & 84.2   & 79.6     & 401.8     & 70.5   & 28.46  \\
OCRNet  & Res-101    & 36.7    & 59.2    & 84.0   & 79.0     & 361.7     & 58.1   & 31.71  \\
ETC   & OCRNet     & 37.5    & 59.1    & 84.1   & 79.1     & 361.7     & 58.1   & -    \\
NetWarp & OCRNet     & 37.5    & 58.9    & 84.0   & 79.0     & 1207.0    & 58.1   & -      \\
MPVSS &Res-101 &38.8 & 59.0 &84.8 & 79.6 &45.1 &103.1 &-\\
Segformer  & MiT-B2     & 43.9    & 63.7    & 86.0   & 81.2     & 100.8     & 24.8   & 30.61  \\
SegFormer & MiT-B5     & 48.2    & 65.1    & 87.8   & 83.7     & 185.0     & 82.1   & 16.82  \\
DCFM($K$=2) &MiT-B2 &43.7 &63.7 &87.7 & 83.2 &22.9 &24.8 &-\\
DCFM($K$=2) &MiT-B5 &48.2 &65.5 &89.0  & 85.0 &57.0 &82.1 &-\\
MRCFA   & MiT-B2     & 45.3    & 64.7    & 90.3   & 86.2     & 127.9     & 27.3   & 23.25  \\
MRCFA    & MIT-B5     & 49.9    & 66.0    & 90.9   & 87.4     & 373.0     & 84.5   & 12.06   \\
CFFM-VSS & MiT-B2     & 44.9    & 64.9    & 89.8   & 85.8     & 143.2     & 26.5   & 22.53  \\
CFFM-VSS  & MiT-B5     & 49.3    & 65.8    & 90.8   & 87.1     & 413.5     & 85.5   & 11.32  \\
SD-CPC(ours) & MiT-B2     & 46.2    & 65.0    & 90.4   & 86.5    &107.1    & 26.0   & 23.86  \\
SD-CPC(ours) & MiT-B5  &{\bf 51.1} & {\bf 66.2}  & {\bf 91.2} & {\bf 87.9} & 324.9 & 83.5  & 12.31 \\
\bottomrule
\end{tabular}
\end{table*}
\noindent
{\bf Dataset.} Our experiment is primarily conducted on the VSPW dataset, which has dense annotations and a high frame rate of 15 FPS, making it the best standard for VSS to date. The VSPW training set, validation set, and test set contain 2,806 clips (198,244 frames), 343 clips (24,502 frames), and 387 clips (28,887 frames), respectively, with a total of 124 categories. In addition, we also evaluate the proposed method on the Cityscapes \cite{cityscapes} dataset, which only has one frame annotation every 30 frames.

\noindent
{\bf Evaluation Metrics.} Following previous work\cite{coarse,mining-relation,motion-state, xie2021segformer, zheng2024deep}, we apply mean IoU (mIoU) and Weight IoU (WIou) to report the segmentation performance. Frames-per-second (FPS), parameters and GFLOPs are used to present the efficiency. Mean video consistency of 8 frames (mVC$_8$) and mean video consistency of 16 frames (mVC$_{16}$) is used to present the video consistency (VC).

\subsection{Comparsions with state-of-the-art Methods} 
The proposed method is compared with state-of-the-art (SOTA) methods on the VSPW dataset, including CFFM-VSS \cite{coarse}, MPVSS \cite{weng2024mask}, MRCFA \cite{mining-relation}, DCFM \cite{zheng2024deep}, SegFormer\cite{xie2021segformer}, OCRNet \cite{OCRNet}, PSPNet \cite{pspnet}, DFF \cite{zhu2017deep}, ETC \cite{ETC}, DVSN \cite{dvsnet}, CC \cite{CC}, GRFP \cite{GRFP} and NetWarp \cite{UperNet_Netwarp}.

In Table \ref{tab:vspw_sota}, we use 20M as the threshold for categorizing model sizes, and separately discuss small and large models. For small models (the first four rows in Table \ref{tab:vspw_sota}), the proposed method demonstrates a 3.4\% increase in mIoU compared to the strong baseline model SegFormer. Additionally, it shows improvements of 4.2\% and 4.8\% in mVC$_8$ and mVC$_{16}$, respectively. Compared to SOTA models MRCFA and CFFM, the proposed method not only significantly improves mIoU and VC, but also reduces computational cost. Specifically, when using MiT-B5, the GFLOPs of the proposed method are reduced by 21.4\% and 12.9\% compared to CFFM-VSS and MRCFA, respectively. It should be noted that lower GFLOPs do not necessarily equate to higher FPS. Similar to EfficientNet \cite{efficientnet}, the proposed method is constrained by GPU bandwidth and requires significant time for data read/write operations, which limits the improvement in FPS. For large models (from the fifth row to the last row in Table \ref{tab:vspw_sota}), the proposed method also outperforms other comparison methods with impressive performance advantages. The results prove the scalability and stability of the proposed method.

In Table\ref{tab:city_sota}, we verify the robustness of the proposed method on the semi-supervised Cityscapes dataset. Our model achieves SOTA results with lower computation costs under two networks of different depths, MiT-B0 and MiT-B1. The results prove that our model effectively captures class-level dependencies and aggregate spatio-temporal information even in a semi-supervised setting.
\begin{figure*}[tb]
  \centering
  \includegraphics[width=\linewidth]{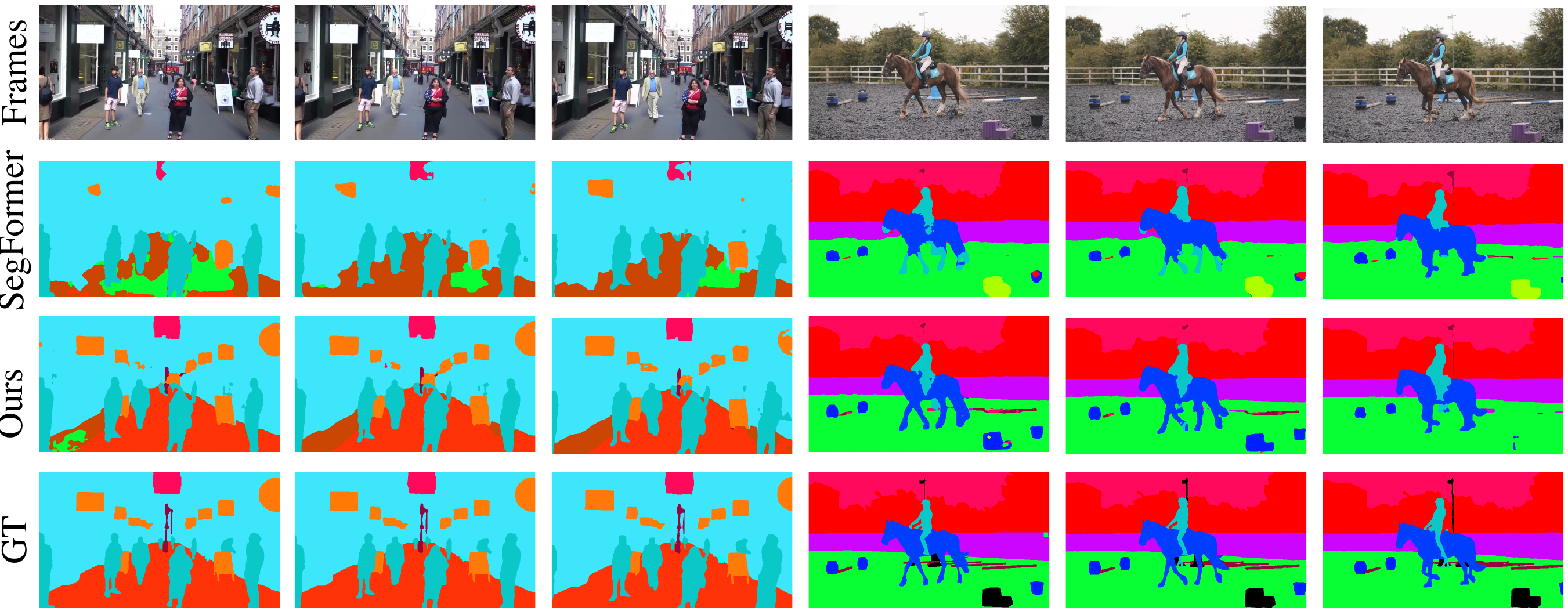}
  \caption{Qualitative results. We compare the proposed method with the baseline (SegFormer with backbone MiT-B1) visually. From top to down: the input video frames, the predictions of SegFormer, our predictions, and the ground truth (GT). The proposed method generates better results than the baseline in terms of accuracy and VC.}
  \label{fig:qualitive}
\end{figure*}

We also qualitatively compare the proposed method with the baseline on the sampled video clips in Table \ref{fig:qualitive}. It is obvious that the proposed scheme can generate more accurate and consistent segmentation results in the complex scenes.
\begin{table}[!t]
\caption{SOTA comparison on the Cityscapes dataset.}
\setlength{\tabcolsep}{1mm}
\label{tab:city_sota}
\begin{tabular}{c|c|c|c|c}
\toprule
Methods & Backbone & mIoU$\uparrow$  & GFLOPs$\downarrow$ &FPS(f/s)$\uparrow$   \\
\midrule
CC  & VGG-16 & 67.7         & -            &  -          \\
GRFP & Res-101   & 69.4       & -       &  -          \\
DVSN & Res-101     & 70.3         & 978.4         &  -          \\
Accel & Res-101     & 72.1         & 824.4             &  - \\
DFF  & Res-101     & 68.7         & 100.8          &  -          \\
ETC & Res-101     & 71.1         & 434.1        & -          \\
\midrule
SegFormer & MiT-B0   & 71.9    & 121.2  & 49.41   \\
MRCFA     & MiT-B0     & 72.8    & 77.5   & 32.81      \\
CFFM-VSS  & MiT-B0    & 74.0     & 80.7    & 31.37      \\
SD-CPC(Ours) & MiT-B0  & {\bf 75.0}  &  49.3   & 29.35      \\
\midrule
SegFormer & MiT-B1   & 71.9  & 121.2   & 49.41  \\
MRCFA  & MiT-B1      & 75.1   & 145.0    & 25.62     \\
CFFM-VSS  & MiT-B1   & 75.1    & 158.7    & 23.54    \\
SD-CPC(Ours)& MiT-B1  & {\bf 76.4}   & 87.4    & 25.80        \\  
\bottomrule
\end{tabular}
\end{table}

\subsection{Ablation Studies} 
We conduct ablation experiments on the VSPW validation set using MiT-B1 as the backbone to demonstrate the effectiveness of each component of our method. All experiments use the same settings as before.
\begin{table}[!t]
\caption{Ablation study on the SD-CPC. “w/o” indicates that the module is removed.}
\setlength{\tabcolsep}{1mm}

\label{tab:abl_SD-CPC}
\centering
\begin{tabular}{c|c c c|c}
\toprule
Methods  & mIoU$\uparrow$  & mVC$_8$$\uparrow$  & mVC$_{16}$$\uparrow$    & GFLOPs$\downarrow$  \\
\midrule
SegFormer   & 36.5    & 84.7   & 79.9     & 26.6 \\
\midrule
\midrule 
SD-CPC & {\bf 39.90} & {\bf 88.9} & {\bf 84.7} & 69.20 \\
\midrule
only DSSA  & 37.94 & 87.6 & 83.2 & 64.25 \\
only SSEA & 37.50 & 87.5 & 82.9 & 49.30 \\
w/o DSSA & 38.52 & 88.3 & 83.7 & 49.30 \\
w/o SSEA & 38.32 & 88.1 & 83.4 & 61.00\\ 
w/o MCP-CL &38.48 &88.4  &83.8  & 69.20 \\
\midrule
w/o stage 1  & 39.04 & 88.5 & 83.9 & 59.25\\
w/o stage 2 & 39.21 & 88.7 & 84.4 & 59.25\\
MCP-CL($M$=1)  & 39.45 & 88.8 & 84.6 & 68.91\\
w/o LA & 38.98 & 88.7 &84.2  & 68.16\\
w/o DCN & 38.51 & 88.4 & 84.1 & 65.61\\
w/o Multi-Scale & 38.84 & 88.2 & 83.5 & 65.34\\
\bottomrule
\end{tabular}
\end{table}

\noindent
{\bf Ablation study on SD-CPC}. 
In Table \ref{tab:abl_SD-CPC}, We first validated the roles of each component of SD-CPC (from the third row to the seventh row), and then verified the necessity of the internal elements of each module (from the eighth row to the last row). When retaining only a single module, our proposed approach still achieves improvements in mIoU by 1.44\% and 1.00\% compared to the baseline model, demonstrating the effectiveness of each module. When removing just one module, any combination of the remaining two modules results in higher mIoU and VC. The results indicates that the three components are tightly coupled and complement each other, contributing to VSS from different aspects. Additionally, the two-stage aggregation improves mIoU by 0.69\% and 0.86\% compared to single-stage aggregation, and multivariate class prototypes enhance mIoU by 0.45\% compared to single-variate class prototypes. This demonstrates that two-stage aggregation and multivariate prototypes capture more spatio-temporal information and category representation capabilities. Finally, we conducted ablation studies on each component of SSEA (the last three rows of the table). The experimental results demonstrate that SSEA, with its simple yet effective design, integrates the strengths of each part to achieve static semantic aggregation with low computation costs (low GFLOPs).

\noindent
{\bf The influence of attention points $P$}. 
The size of $P$ should be adjusted according to different scenarios to strike a balance between segmentation performance and computation cost. During the experiments, as $P$ increased from 4 to 9, segmentation performance and temporal consistency experienced improvements (0.2 in mIoU, 0.1 in mVC$_{8}$, 0.2 in mVC$_{16}$), while FPS decreased from 32.54 to 27.25. With a further increase in $P$ to 16, segmentation performance continued to improve (0.5 in mIoU, 0.3 in mVC$_8$, 0.5 in mVC$_{16}$), while FPS decreased from 32.54 to 21.13. The experimental results are reasonable because as $P$ increases, cross-frame selective cross-attention mechanism gradually approximates vanilla cross-attention mechanism, gaining more information while also increasing computation cost.

\section{Conclusion}
In this paper, we rethink the static and dynamic contexts in VSS from the perspective of class-level perceptual consistency and propose a novel SD-CPC framework. Specifically, we introduce a multivariate class prototype with contrastive learning to impose class-level constraints. And, we propose a static-dynamic semantic alignment module, whereby the static semantics provide a reliable foundation for the selective aggregation of dynamic semantics, and the dynamic semantics leverage the inter-frame associations to enhance the static semantics. To avoid redundant computations, we propose a window-based attention map calculation method that leverages the sparsity of attention points, thereby reducing the computational complexity. Extensive experiments demonstrate that the proposed method  significantly outperforms the current state-of-the-art approaches, showing great potential for exploring VSS through static-dynamic class-level perceptual consistency.

\bibliography{main}
\end{document}